%% file: stt.tex
\documentclass[letterpaper, 10 pt, conference]{ieeeconf}  % Comment this line out if you need a4paper

\usepackage{times}
\usepackage{epsfig}
\usepackage{graphicx}
\usepackage{amsmath}
\usepackage{amssymb}
\usepackage{booktabs}
\usepackage{multirow}
\usepackage{float}
\usepackage{color}
\usepackage{arydshln}
\usepackage[export]{adjustbox}
\usepackage{pifont}
\usepackage{booktabs,subcaption,amsfonts,dcolumn}
\newcommand{\captionvspace}{0mm}%

\newcommand{\comment}[1]{}%

\usepackage[capitalize]{cleveref}
\crefname{section}{Sec.}{Secs.}
\Crefname{section}{Section}{Sections}
\Crefname{table}{Table}{Tables}
\crefname{table}{Tab.}{Tabs.}

\usepackage{mathtools}
\usepackage{multibib}

\IEEEoverridecommandlockouts                              % This command is only needed if 
\title{\LARGE \bf
STT: Stateful Tracking with Transformers for Autonomous Driving
}

\author{Longlong Jing$^{*}$, Ruichi Yu$^{*\dagger}$, Xu Chen$^{*}$, Zhengli Zhao, Shiwei Sheng, \\ 
Colin Graber, Qi Chen, Qinru Li, Shangxuan Wu, Han Deng, Sangjin Lee, \\ 
Chris Sweeney, Qiurui He, Wei-Chih Hung, Tong He, Xingyi Zhou$\ddagger$, \\ 
Farshid Moussavi, James Guo, Yin Zhou, Mingxing Tan, Weilong Yang, Congcong Li \\
Waymo LLC, $\ddagger$Google Research
\thanks{$^{*}$Equal Contributions. }%
\thanks{$\dagger$Corresponding author.}%
}

\begin{document}
\maketitle
\thispagestyle{empty}
\pagestyle{empty}

%%%%%%%%%%%%%%%%%%%%%%%%%%%%%%%%%%%%%%%%%%%%%%%%%%%%%%%%%%%%%%%%%%%%%%%%%%%%%%%%
\begin{abstract}

Tracking objects in three-dimensional space is critical for autonomous driving. To ensure safety while driving, the tracker must be able to reliably track objects across frames and accurately estimate their states such as velocity and acceleration in the present. Existing works frequently focus on the association task while either neglecting the model's performance on state estimation or deploying complex heuristics to predict the states. In this paper, we propose STT, a \underline{\emph{S}}tateful \underline{\emph{T}}racking model built with \underline{\emph{T}}ransformers, that can consistently track objects in the scenes while also predicting their states accurately. STT consumes rich appearance, geometry, and motion signals through long term history of detections and is jointly optimized for both data association and state estimation tasks. Since the standard tracking metrics like MOTA and MOTP do not capture the combined performance of the two tasks in the wider spectrum of object states, we extend them with new metrics called S-MOTA and MOTP$_\text{S}$ that address this limitation. STT achieves competitive real-time performance on the Waymo Open Dataset.

\end{abstract}

%%%%%%%%% BODY TEXT
\input{introduction}
\input{related_work}
\input{method}
\input{experiments}
\input{conclusion}

\input{acknowledgement}

{\small
\bibliographystyle{IEEEtran}
\bibliography{egbib}
}

\end{document}

%% file: introduction.tex
\section{Introduction}
\label{sec:intro}

3D Multi-Object Tracking (3D MOT) plays a pivotal role in various robotics applications such as autonomous vehicles. To avoid collisions while driving, robotic cars must reliably track objects on the road and accurately estimate their motion states, such as speed and acceleration. While development of 3D MOT has made much progress in recent years, most methods~\cite{pang2021simpletrack, Weng2019_3dmot, wang2021immortal} still use approximated object states as intermediate features for data association without explicitly optimizing model performance on state estimation. Although some tracking methods~\cite{tracker_robotic1, tracker_robotic2, tracker_robotic3, tracker_robotic4} exist that predict motion states, they often do so by employing filter-based algorithms such as the Kalman filter (KF) with complex heuristic rules~\cite{pang2021simpletrack, wang2021immortal, yin2021center} to estimate object states and cannot easily utilize appearance features or raw sensor measurements in a data-driven fashion~\cite{xiang2015learning}. While there are machine learning-based methods~\cite{zhou2020tracking} that add prediction heads to detection models to estimate motion states, they struggle to produce consistent tracks from long-term temporal information due to computational and memory limitations.

To address the limitations of existing approaches, we introduce STT, a \underline{S}tateful \underline{T}racking model with \underline{T}ransformers, which combines data association and state estimation into a single model. At the core of our model architecture are a Track-Detection Interaction (TDI) module that performs data association by learning the interaction between a track and its surrounding detections and a Track State Decoder (TSD) that produces the state estimation of the tracks. 

All the modules are jointly optimized (Figure~\ref{fig:framework}), which allows STT to obtain superior performance while simplifying the system complexity.

Existing tracking evaluation mainly use multi-object tracking accuracy (MOTA) and multi-object tracking precision (MOTP)~\cite{bernardin2006multiple} to measure the association and localization quality, but they do not take the quality of other states into account such as velocity and acceleration. To explicitly capture the full state estimation quality of the tracking performance, we extend the existing evaluation metric MOTA to Stateful MOTA (S-MOTA) which enforces accurate state estimation during label-prediction matching, and MOTP to MOTP$_\text{S}$ which applies to arbitrary state variables so that we can assess the quality of the state estimation beyond position.

To demonstrate the effectiveness of our STT model, we conduct extensive experiments on the large-scale Waymo Open Dataset (WOD)~\cite{sun2020scalability}. Our model achieves competitive performance with 58.2 MOTA and state-of-the-art results in our extended S-MOTA and MOTP$_\text{S}$ metrics. We conduct comprehensive ablation studies for STT, which allows us to better understand its performance. 

The contributions of this work are summarized as follows: {
\begin{enumerate}
  \item We propose a 3D MOT tracker which tracks objects and estimates their motion states in a single trainable model.
  \item We extend the existing evaluation metrics to S-MOTA and MOTP$_\text{S}$ to evaluate tracking performance that explicitly considers the quality of the state estimation.
  \item Our proposed model achieves improved performance over strong baselines with standard metrics and state-of-the-art results with the newly extended metrics on the Waymo Open Dataset.
\end{enumerate}
}

\begin{figure}
\begin{center}
\includegraphics[width=0.7\linewidth]{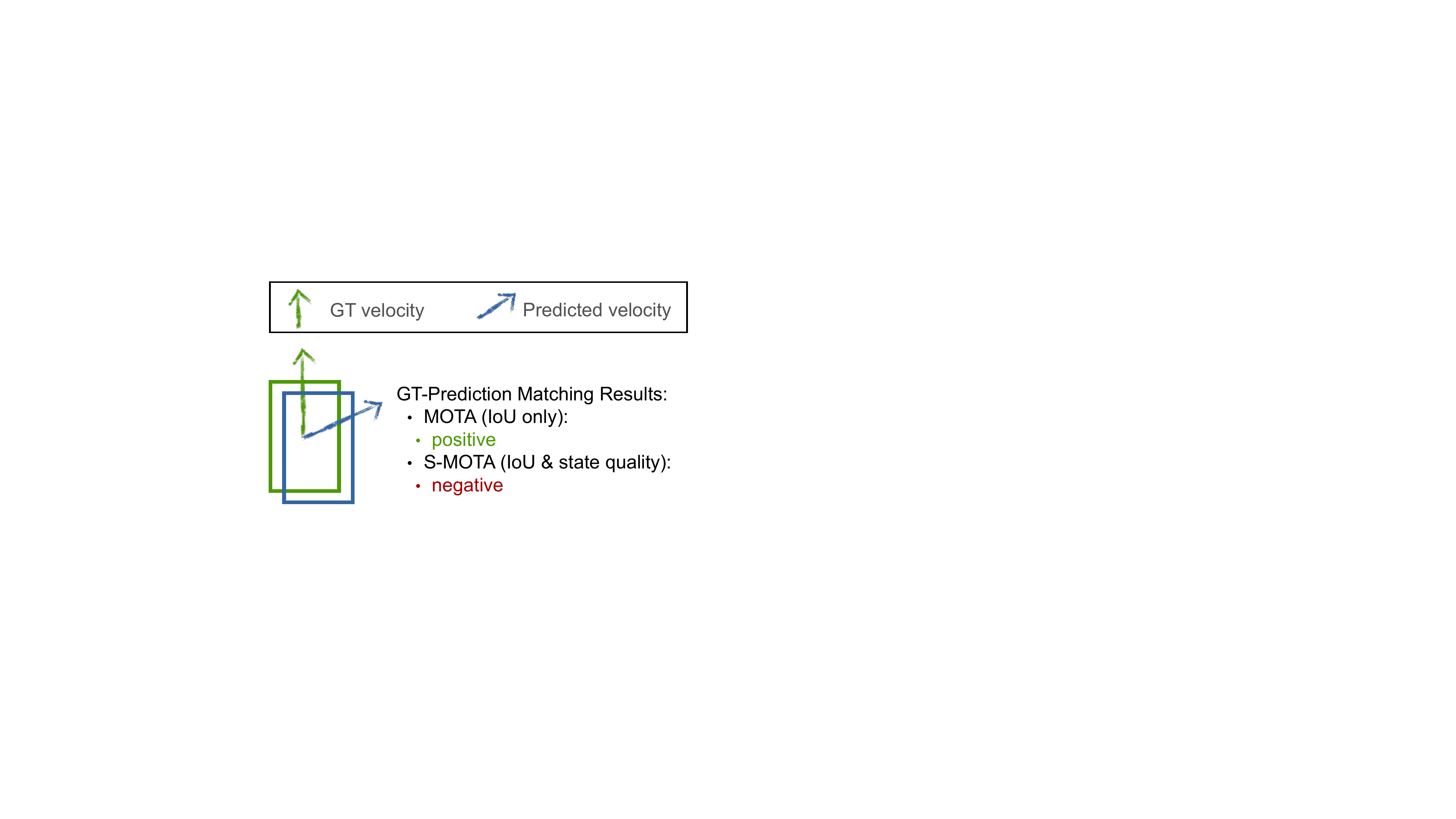}
\end{center}
\vspace{\captionvspace}
\caption{
\textbf{Illustration of S-MOTA metric.}
MOTA~\cite{MOT16} only considers IoUs in label-prediction matching, and does not reveal state errors (e.g., velocity error shown in the figure). This limitation is addressed by S-MOTA via an additional thresholding step to assess the accuracy of predicted state.
}
\label{fig:metrics}
\vspace{-15pt}
\end{figure}

%% file: related_work.tex
\section{Related Work}

\noindent\textbf{2D Multi-Object Tracking}~\cite{MOTChallenge2015,MOT16,geiger2012we}
aims to track objects in crowd scenes~\cite{chu2021transmot,peng2020tpm,peng2020chained,Wu2021TraDeS,yu2007multiple,wang2019towards,dai2021learning,zeng2021motr,xu2021transcenter,qdtrack,transtrack,wang2021multiple,zhou2022global, xu2019spatial, xiang2023hm, zhou2020tracking, meinhardt2021trackformer}, and the dominant methods follow a tracking-by-detection paradigm~\cite{Bewley2016_sort,Wojke2017simple,bergmann2019tracking,tang2017multiple,zhang2020fair}. 2D MOT approaches rarely estimate the motion state of objects since it is challenging to perform 3D state estimation from 2D data and the motion states estimated from a perspective view are often not informative for downstream modules in autonomous driving.

\noindent\textbf{3D Multi-Object Tracking} is a popular problem in autonomous driving~\cite{zhou2022geometry,kim2022polarmot,gladkova2022directtracker,kim2021eagermot, hung2020soda, xu2022opv2v}. Compared to 2D tracking, this problem space is less explored. Prior works in 3D tracking have primarily relied on Kalman Filters~\cite{Weng2019_3dmot,kuang2020probabilistic,wang2021immortal}, as seen in numerous state-of-the-art methods on the Waymo Open Dataset. Other works explore learning-based solutions~\cite{pang2021quasi,hu2022monocular}. Unlike these works that either ignore or separate the state estimation task from association task, our STT model can learn these two tasks together.

\noindent\textbf{State Estimation} is a problem domain where the goal is to predict the state of an object including its dynamic attributes (e.g., speed, acceleration) and semantic attributes (e.g., object type, appearance). Existing tracking solutions primarily focus on the dynamic attributes for state estimation, as these are highly correlated with tracking performance. Common practices include predicting them using a motion filter that smooths estimations over time~\cite{Weng2019_3dmot,wang2021immortal} and including them as an output in an object detection model~\cite{zhou2020tracking,chen2023voxelnext}. Compared to these methods, our approach has a dedicated machine learning module that can encode the temporal features from a detection model and predict accurate object state.

\noindent In \textbf{Multi-Object Tracking Evaluation}, the most commonly used metric~\cite{sun2020scalability,nuscenes2019} is the MOTA~\cite{bernardin2006multiple,MOT16}. It captures both the detection box quality and tracking performance. However, it only explicitly evaluates the position result and does not directly evaluate other object states. MOTP~\cite{bernardin2006multiple} also only considers the localization error of the positive matches in MOTA. The stateful metrics we propose consider a wider range of state estimates jointly with association, and thus better reflect the overall tracking quality. While MOTA can be combined with other standalone metrics for assessing the state estimation~\cite{nuscenes2019}, S-MOTA uses a single unified metric that highlights the estimation quality across all states and MOTP$_\text{S}$ offers fine-grained evaluation on any generic state. Other tracking metrics like IDF1~\cite{stiefelhagen2006clear} and HOTA~\cite{luiten2021hota} put more emphasis on data association quality and are complementary to our proposed metrics.

\label{sec:related}

%% file: method.tex
\section{Methodology}
\label{sec:method}

\begin{figure*}
\begin{center}
\includegraphics[width=0.8\linewidth]{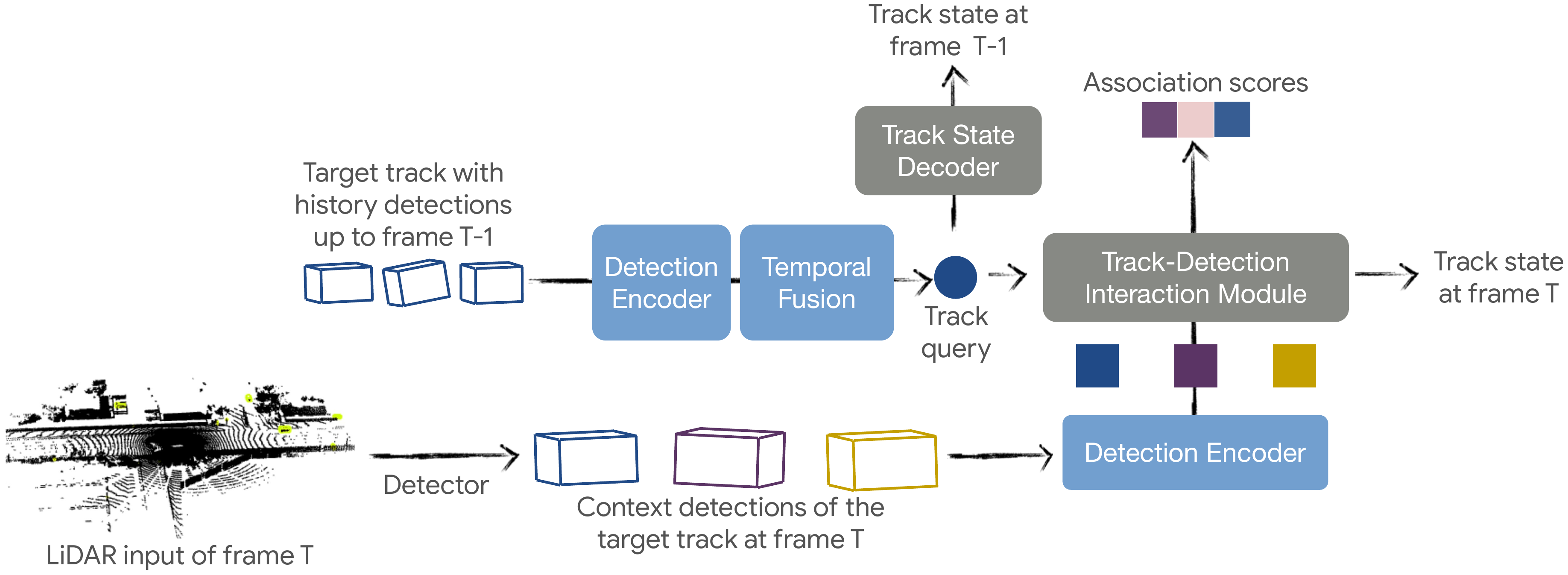}
\end{center}
\vspace{-10pt}
\caption{\textbf{Overview of STT.} We first use the Detection Encoder to encode all of the 3D detections and extract temporal features for each track. The temporal features are fed into the Track-Detection Interaction module to aggregate information from surrounding detections and produce association scores and predicted states for each track. The Track State Decoder also takes the temporal features to produce track states in the previous frame $t-1$. All modules are jointly optimized.}
\label{fig:framework}
\vspace{-15pt}
\end{figure*}

In this section, we will first formalize the tracking problem and then describe the architecture of our STT model. We will cover its training and inference process and discuss our new tracking metrics that cover a wide spectrum of the object states. An overview of STT is shown in Figure~\ref{fig:framework}.

% =====================================
\subsection{The Tracking Problem}
\label{sec:tracking_problem}

The goal of the tracking problem discussed in this paper is to maintain a set of tracks $\vec \tau_1^t, \vec \tau_2^t, \ldots, \vec \tau^t_{N^t}$ for the $N^t$ objects in a scene at time $t$, where each tracklet $\vec \tau_n^t = [S_n^{t_{k}}, \ldots, S_n^t]$ consists of a list of state vectors $S_n^t$ from $t_{k}$ to the current time $t$. The state vector $S_n^t$ is defined as $S_n^t = [\{s\}|_{s \in \mathcal{S}}]$, where $s \in \mathbb{R}^{d_s}$ is a $d_s$-dimensional vector representing state type $s$, $\mathcal{S}$ is the set of state types being considered, and $[\cdot]$ is the concatenation operation. In this work, we model states $S_n^t = [\mathbf{x}, \mathbf{v}, \mathbf{a}] \in \mathbb{R}^6$, i.e., the concatenation of position $\mathbf{x} \in \mathbb{R}^2$, velocity $\mathbf{v}\in \mathbb{R}^2$, and acceleration $\mathbf{a}\in \mathbb{R}^2$. Each state type is defined over the $XY$ plane, as objects on the road rarely move alone the $Z$ direction. Nevertheless, the problem can be easily generalized to the $Z$ direction.

Assume that the tracks are given as $\vec \tau_1^{t-1}, \vec \tau_2^{t-1}, \ldots, \vec \tau^{t-1}_{N^{t-1}}$ at time $t-1$, and a new set of 3D detection are given at time $t$ as $p_1, p_2, \ldots, p_{N^t}$, where $p_i = (b_i, o_i, f_i)$ with bounding box $b_i$, appearance features $o_i$, and confidence score $f_i \in [0, 1]$. The box $b_i \in \mathbb{R}^7$ contains the position $(x, y, z)$, sizes (width, length, height), and heading. The tracking problem is then defined as computing the tracks $\vec \tau_1^t, \ldots, \vec \tau^t_{N^t}$ and their states $S_1^t, \ldots, S_{N^t}^t$ at time $t$. Note that $N^t$ can be different from $N^{t-1}$, as new tracks can be created and the existing tracks can be deleted due to the lack of observations. 

% =====================================
\subsection{Modeling}

\subsubsection{Detection Encoder and Temporal Fusion}
As a tracking model, STT can interact with arbitrary 3D detection models. To ensure that STT can learn a descriptive embedding that captures the geomtry, appearance, and motion features of the detection, we design a Detection Encoder (DE) to encode the detection outputs:

\begin{equation}
\text{emb}(\text{det}_i) = \text{DE}(g_i, a_i, m_i, \mathbf{\theta}_{\text{DE}})
\end{equation} 
Let $\text{det}_i$ denote the $i$th detection, and let $g_i, a_i, m_i$ be the corresponding geometry, appearance, and motion features for this detection respectively. $\mathbf{\theta}_{\text{DE}}$ are the learned parameters of DE. DE is implemented as a multilayer perceptron (MLP) in our model.

After the DE comes a Temporal Fusion (TF) model that combines these detection embeddings over time to create a temporal embedding that describes each track's history. To better model the historical context of a track $\vec \tau_j^{t-1}$, we apply a self-attention model to the associated detection embeddings and obtain the track query $Q_{\vec \tau_j^{t-1}}$ at time $t-1$:
\begin{equation} \label{eq1}
\begin{split}
 Q_{\vec \tau_j^{t-1}}  = \text{TF}(\{\text{emb}(\text{det}_i)|i=1,...,t-1\}, \mathbf{\theta}_{\text{TF}})
\end{split}
\end{equation}
where $\text{det}_i \in \mathbf{Det}(\vec \tau_j^{t-1})$, and $\mathbf{Det}(\vec \tau_j^{t-1})$ is the set of associated detections for track $\vec \tau_j^{t-1}$ until time $t-1$. After self-attention, TF aggregates the embeddings $\mathbb{R}^{1\times T \times D_q}$ across time and outputs the self-attended embedding in $\mathbb{R}^{1 \times D_q}$ at time $t-1$. $T$ is the track length, $D_q$ is the feature size, and $\mathbf{\theta}_{\text{TF}}$ are the learned parameters.

\subsubsection{Track State Decoder}
For a track $\vec \tau_j^{t-1}$ at time $t$, the track query $Q_{\vec \tau_j^{t-1}}$ %already
encodes its history up to time $t-1$. Therefore, we can directly predict the state $\mathbf{S}_{t-1}$ for every track with a light-weight Track State Decoder (TSD) module:
\begin{equation}
\textbf{S}_{t-1} = G(\mathbf{Q}_{t-1}, \mathbf{\theta_g})
\end{equation}
where $\mathbf{Q}_{t-1}$ is the list of all the track queries. $G$ is a MLP and $\mathbf{\theta_g}$ are its learned parameters. TSD helps us supervise the track embedding, but it is also useful as a stand-alone state estimator for a given track embedding at any given timestamp. We will elaborate more on how this decoder is used during a typical tracker update loop in Section~\ref{sec:method:OTI}.

\subsubsection{Track-Detection Interaction Module}
The Track-Detection Interaction (TDI) module calculates the relationship between tracks and their surrounding context detections at time $t$. For each track $\vec \tau_j^{t-1}$ from time $t-1$, we select $k$ context detections $\mathbf{K}_n$ from all the detections $\mathbf{M}$ at time $t$ in a small area around the track:
\begin{equation}
\mathbf{K}_n = \{b_i |D(\text{pred}(\vec \tau_j^{t-1}), b_i) < d, b_i \in p_i, p_i \in \mathbf{M}\}
\end{equation} 
where $D$ computes the distance between detection $b_i$ and the track's state estimation $\text{pred}(\vec \tau_j^{t-1})$ at time $t$. During training, we directly use the ground truth state at time $t$ to represent $\text{pred}(\vec \tau_j^{t-1})$. During inference, we extrapolate the estimated track state at time $t-1$ to time $t$ to search for the context detections effectively before running the model. In practice, we set threshold $d$ to be small enough for efficiency, but large enough to ensure that all the detections of true positive association for track $\vec \tau_j^{t-1}$ are included in the context set $\mathbf{K}_n$.

We use the same Detection Encoder to create the detection embeddings $\mathbf{C_i}$ in $\mathbf{K}_n$. The TDI module then takes the list of queries $\mathbf{Q}_{t}$ and $\mathbf{C_i}$ as input to predict the association scores for all the tracks and detections:
\begin{equation}
\mathbf{AS} = \text{TDI}(\mathbf{Q}_{t}, \mathbf{C_i}, \mathbf{\theta}_{\text{TDI}})
\end{equation} 
where $\mathbf{\theta}_{\text{TDI}}$ are learned parameters. $\mathbf{AS}=\{AS\}$, where $AS \in \mathbb{R}^{1\times k}$ are the association scores between a track query $Q_{\vec \tau_j^{t-1}}$ and the $k$ context detections. TDI is a transformer-based model~\cite{vaswani2017attention} with an added MLP to predict the track state at time $t$ after cross-attending to the context detections.

% =====================================
\subsection{Training}
Our model is jointly trained using a data association loss $L_d^t$ and state estimation losses $L_s^{t}$, $L_s^{t-1}$:
\begin{equation}
L_{\text{total}} = \gamma L_d^t + \lambda L_s^{t} + \alpha L_s^{t-1}
\end{equation}
where $\gamma$, $\lambda$, and $\alpha$ are the weight of each loss term. We optimize the per-track query with per box association loss. Let $AS_i$ be the association score between the track query $Q_{\vec \tau_j^{t-1}}$ and one of its context detections $\text{det}_i$. And let $y$ be the ground-truth association with 0 as ``not associated'' or 1 as ``association'' Then the loss of this pair is:
\begin{equation}
L(Q_{\vec \tau_j^{t-1}}, \text{det}_i) = -{(y\log(AS_i) + (1 - y)\log(1 - AS_i))}
\end{equation}
For each track query, the total association loss is computed against all of its context detections as:
\begin{equation}
L_d^t = \sum_{i=1}^{k} L(Q_{\vec \tau_j^{t-1}}, \text{det}_i)
\end{equation}
where $k$ is the number of context detections.

The state estimation losses are the L1 loss between the predicted states and the ground truth states for each track at time $t$ (via the output of TDI module) and $t-1$ (via the output of the TSD module): 
\begin{equation}
L_s^t =  \left|\textbf{S}_j^t - \textbf{S}_j^{*t} \right|, L_s^{t-1} = \left|\textbf{S}_j^{t-1} - \textbf{S}_j^{*t-1} \right|
\end{equation} 
where $S_j^{*t}$ and $S_j^{*t-1}$ is the ground truth state for the track $\vec \tau_j^{t}$ and $\vec \tau_j^{t-1}$ respectively. 

% =====================================
\subsection{Online Tracker Inference}
\label{sec:method:OTI}
During tracking inference, we apply STT over the laser stream frame by frame. For each frame at time $t$, a 3D object detection model is first applied over the laser spin to get all $N$ detection boxes. For each detection box, its geometry features, appearance features, and confidence score are collected as ${p_n^t}$, while $p^t$ is the list of all the detections' feature vectors. For all tracks produced from the previous frame at time $t-1$, we cache their learned track query $\mathbf{Q}_{t-1}$. Then, the TDI module is applied over the queries $\mathbf{Q}_{t-1}$ and all detection embeddings $\mathbf{emb}(p^t)$ to produce the association likelihood 2D matrix $\mathbf{AS}$ between all the tracks and boxes.

The Hungarian matching algorithm~\cite{kuhn1955hungarian} is then applied over $\mathbf{AS}$ to produce the assignment result. If the association score is lower than a pre-defined threshold, a new track will be created. Otherwise, the detection will be assigned to an existing track query and appended to its history. For the first frame of a track, all the detected boxes are treated as new tracks and their initial states (e.g. velocity and acceleration) will be set to $0$. For all the subsequent frames, we use TSD to predict state for the track at time $t$ as we find that it is slightly better than the output of TDI.

% =====================================
\subsection{Stateful Evaluation Metrics}
\subsubsection{S-MOTA}
MOTA~\cite{bernardin2006multiple} is one of the most commonly used metrics for multiple object tracking. Computing MOTA involves a matching step similar to the evaluation of object detection. A given prediction-label pair $(p, g)$ is only considered for matching if their IoU is larger than a given threshold:
\begin{equation}
    C(p, g) = \begin{cases}
        1 - U(p, g), & \text{if $U(p, g) > t_u$} \\ +\infty, &\text{otherwise}
    \end{cases}
\end{equation}
$U(\cdot)$ is the IoU function and $t_u$ is a class-specific threshold. $C(\cdot)$ denotes the cost function of the matching algorithm. Consequently, MOTA primarily evaluates the quality of the detections as well as the predicted associations. The only component of the states defined in Section~\ref{sec:tracking_problem} evaluated here is the location (i.e., the detection box center), and the prediction accuracies of other states are only indirectly evaluated through the improvements they may bring to association. 

To better evaluate data association and state estimation, we extend the MOTA to \emph{Stateful Multiple Object Tracking Accuracy} (S-MOTA). This is computed using the same procedure as standard MOTA, but with additional requirements in the state estimation for a given prediction-label pair to be matched. Accurate state estimation such as a vehicle's velocity is critical for autonomous driving. In S-MOTA, the state estimation error of each pair must be below a class- and state-dependent threshold to allow matching:
\begin{equation}
    C(p, g) = \begin{cases}
        1 - U(p, g), & \parbox{1.2in}{if $U(p, g) > t_u$ and $\cap_{s \in \mathcal{S}} \|p_s - g_s\| < t_{u,s}$} \\
        +\infty, &\text{otherwise}
    \end{cases}
\end{equation}
Let $p_s$ and $g_s$ denote predicted/ground-truth state vectors of type $s$. $\mathcal{S}$ is the set of states considered for the evaluation, and $t_{u,s}$ is the threshold for state type $s$ and class $u$. Hence, maximizing S-MOTA requires track predictions to both have proper associations across time as well as reasonably close state predictions. For this work, $\mathcal{S}$ consists of velocity and acceleration. In principle, however, any combination of state types from a tracker can be used to derive a S-MOTA metric.

\subsubsection{MOTP$_\text{S}$}
The extended S-MOTA metric is designed to provide a comprehensive evaluation of tracking performance, including state estimation. As a complement, we extend the MOTP to Multiple Object Tracking Precision for General States (MOTP$_\text{S}$) to provide more fine-grained evaluation on the state estimation accuracy. Given the set $\mathcal{M}$ containing pairs of predictions $p$ and label $g$ which are matched during MOTA computation, MOTP$_\text{S}$ computes the average L2 error for each state type to measure the magnitude of the state error, i.e., for each state type $s \in \mathcal{S}^*$:
\begin{equation}
   \text{MOTP}_s(\mathcal{M}) = \tfrac{1}{|\mathcal{M}|}\sum_{(p, g) \in \mathcal{M}}\|p_s - g_s\|
\end{equation}

We can further measure the count of objects with large state estimation errors, i.e.,
\begin{equation}
    \left|\text{MOTP}_s(\mathcal{M})\right| = \left|\{(p, g) \in \mathcal{M} ~~| ~~\|p_s - g_s\| > \alpha_s\}\right|
\end{equation}
where $\alpha_s$ is a threshold for state $s$. Note that MOTP$_\text{S}$ is consistent with the definition of MOTP. In fact, the latter is a specific version of the former in the localization state. Rather than defining a single metric that aggregates across states, we use separate MOTP$_\text{S}$ metrics for each state type to highlight the performance of each type of state individually.

The evaluation dataset has a disproportionate amount of stationary objects. To ensure that the metrics properly evaluate performance on objects with different types of motion, we report the L2 state error in three different speed breakdowns: static, slow moving objects, and fast moving objects. We also count the number of predictions with L2 error larger than the threshold $\alpha_s$ to focus on challenging cases where the predictions are off significantly.

%% file: experiments.tex
\section{Experiments}
\label{sec:experiments}

\begin{table*}[t!]
\vspace{3pt}
\centering
\caption{Comparison with state-of-the-art tracking methods on the validation set of Waymo Open Dataset.}
\vspace{-5pt}

\resizebox{0.95\textwidth}{!}{
\begin{tabular}{@{}l@{}c@{\ \ \ \ }c@{\ \ \ \ }c@{\ \ \ \ }c@{\ \ \ \ }c@{\ \ \ \ }c@{\ \ \ \ }c@{\ \ \ \ }c@{\ \ \ \ }c@{\ \ \ \ }c@{}} 
 \toprule
 \multirow{2}{*}{Method} & \multicolumn{5}{c}{Vehicle} & \multicolumn{5}{c}{Pedestrian} \\ 
 \cmidrule(lr){2-6} \cmidrule(lr){7-11} &
 S-MOTA$\uparrow$ & MOTA$\uparrow$ &FP$\downarrow$ &Miss$\downarrow$ &Missmatch$\downarrow$ &S-MOTA$\uparrow$ &MOTA$\uparrow$ &FP$\downarrow$ &Miss$\downarrow$ &Missmatch$\downarrow$ \\
 \midrule
 CenterPoint~\cite{yin2021center} & - & 55.1 & 10.8 & 33.9 &0.26 & - &54.9 & 10.0 &34.0 &1.13 \\ 
 SimpleTrack~\cite{pang2021simpletrack} & - & 56.1 & 10.4 & 33.4 &0.08 & - &57.8 &10.9 &30.9 &0.42 \\
 CenterPoint++~\cite{yin2021center} & - & 56.1 & \textbf{10.2} & 33.5 &0.25 & - &57.4 &11.1 &30.6 &0.94 \\
 Immortal Tracker~\cite{wang2021immortal} & - & 56.4 & \textbf{10.2} & 33.4 &\textbf{0.01} & - &58.2 &11.3 &30.5 &\textbf{0.26} \\ 
 Kalman Filter (Ours) & 34.6  & 56.5 & 10.6 & 32.8 & 0.1 & 41.8  & 59.7 & 10.1 & \textbf{29.6} & 0.5 \\
 STT (Ours) & \textbf{48.0} & 58.2 & 10.4 & 31.3 &0.1 & \textbf{55.2} & 59.9 & 10.2 & \textbf{29.6} & 0.3\\
 \midrule TrajectoryFormer~\cite{chen2023trajectoryformer} & - & \textbf{59.7} & 11.7 & \textbf{28.4} & 0.19 & - & \textbf{61.0} & \textbf{8.8} & 29.8 & 0.37 \\
 \bottomrule
\end{tabular}
}
\vspace{-10pt}
\label{table:association-val}
\end{table*}

\begin{table*}[t]
\centering
\caption{Comparisons for MOTP$_\text{S}$ on the validation set of Waymo Open Dataset.}

\vspace{-5pt}
\resizebox{0.95\textwidth}{!}{
\begin{tabular}{@{}l@{}c@{\ \ }c c c c c c c c c c@{}c@{}c@{}} 
 \toprule
 \multirow{3}{*}{Method} & \multirow{3}{*}{Class} &\multicolumn{4}{c}{MOTP$_\text{velocity}$$\downarrow$} &\multirow{3}{*}{$\left|\text{MOTP}_\text{velocity}\right|$$\downarrow$} &\multicolumn{4}{c}{MOTP$_\text{acceleration}$$\downarrow$} &\multirow{3}{*}{$\left|\text{MOTP}_\text{acceleration}\right|$$\downarrow$} \\
 \cmidrule(lr){3-6} \cmidrule(lr){8-11} 
  & & Static & Slow & Fast  &All &  & Static & Slow & Fast &All & \\
 \midrule
  SWFormer\cite{Sun2022SWFormerSW}+SH &\multirow{3}{*}{Vehicle} & \textbf{0.016} & 0.258 & 0.372 & 0.098 & 3063 & \textbf{0.013} & 0.864 & 0.758 & 0.179 & 11089\\
  Kalman Filter & & 0.117 & 0.271 &  0.260 &0.176  & 1890 & 0.217 & 0.683 & 0.665 &0.418 & 25050  \\
STT & & 0.049  &\textbf{0.214}  &\textbf{0.235} &\textbf{0.095}  & \textbf{794}  & 0.026  &\textbf{0.425}  &\textbf{0.412} &\textbf{0.116} & \textbf{1528} \\
  \midrule
  SWFormer\cite{Sun2022SWFormerSW}+SH &\multirow{3}{*}{Pedestrian} & \textbf{0.061} & 0.179 & 0.307 & 0.162 & 147 & \textbf{0.066} & \textbf{0.155} & 0.340 & \textbf{0.135} & 121 \\
 Kalman Filter & & 0.116 & 0.15 & \textbf{0.183} &0.149 & \textbf{25} & 0.212 & 0.345 & 0.422 &0.336 & 6930  \\
STT & & 0.066 &\textbf{0.112} & 0.205 &\textbf{0.100} & 39 & 0.082 &\textbf{0.155} &\textbf{0.324} & 0.141 & \textbf{27} \\
 \bottomrule
\end{tabular}
}
\label{table:state}
\vspace{-20pt}
\end{table*}

\noindent\textbf{Datasets.}
We evaluate our STT model on the Waymo Open Dataset~\cite{sun2020scalability}, which contains $798$ sequences for training, $202$ sequences for validation, and $150$ sequences for testing. Each sequence lasts $20$ seconds at $10$ Hz. Following other popular methods, we evaluate our method on vehicles and pedestrians for the LEVEL 2 difficulty setting~\cite{sun2020scalability}, which is more diffcult than LEVEL 1 because it includes objects with fewer than five laser points in their boxes. LEVEL 2 also includes all the objects in LEVEL 1.

\noindent\textbf{Training details.} Our model is jointly trained on $16$ TPUs with a batch size of $512$. The AdamW~\cite{loshchilov2017decoupled} optimizer is used with $0.03$ weight decay. The initial learning rate is $0.0001$ with linear learning rate decay of $0.5$. The model is trained for $125,000$ steps, including $1,000$ warm-up steps. We set association loss weight $\gamma=10$ and we have different loss weights for different states: $1$ for both position and velocity and $10$ for acceleration. Unless explicitly specified, we set the maximum track length $T=10$ for encoding track history and select a maximum of $20$ context detections for training the model. We use SWFormer~\cite{Sun2022SWFormerSW} as our detection backbone.

\subsection{Overall Results} 

To demonstrate the effectiveness of our STT model, we compare it with published state-of-the-art methods on the Waymo Open Dataset. The majority of the 3D MOT algorithms adopt the tracking-by-detection paradigm, and each of them uses different detection backbones for their tracking algorithms~\cite{pang2021simpletrack, wang2021immortal, yin2021center, chen2023trajectoryformer, li2022time3d, weng20203d}. As STT is a stateful tracker that can be used with arbitrary detection models, we need to compare it with a tracking method that uses the same detection model as STT. Following~\cite{sun2020scalability, Weng2019_3dmot,pang2021simpletrack}, we develop a Kalman Filter baseline that uses the same detection backbone as STT.

We first compare our model with these state-of-the-art methods as well as our KF baseline on the official 3D tracking metrics of the Waymo Open Dataset. These metrics includes MOTA, MOTP, False Positives (FP), False Negatives (FN), and mismatches (Identity Switches). The results are shown in Table~\ref{table:association-val}. Our KF baseline, which uses a strong detection backbone~\cite{Sun2022SWFormerSW}, already achieves competitive performance compared with other existing methods. STT achieves a MOTA score that is +1.7 higher than our KF baseline on the vehicle type and on-par results on other metrics, demonstrating the benefit of including state estimation into the learning process of our tracking model. Note that the miss rate of the KF and STT models are slightly different due to the different cut-off scores used by the two methods. The strong performance of the KF baseline also indicates that these official metrics heavily rely on the quality of the detections. A simple tracker can achieve better performance than other highly-tuned approaches by using a stronger object detector (e.g. our KF baseline vs. CenterPoint~\cite{yin2021center}).

To demonstrate STT's advantage on state estimation over the KF baseline, we further compare them using the stateful metric S-MOTA, as shown in Table~\ref{table:association-val}. This metric requires prediction/ground-truth matches to have sufficiently high predicted velocity and acceleration quality. The velocity and acceleration thresholds are set to $1.0$ m/s and $1.0$ m/s$^2$ for vehicles and $0.5$ m/s and $0.5$ m/s$^2$ for pedestrians. The S-MOTA score of STT is $13.4$ higher than the KF baseline for both vehicles and pedestrians. This shows that while STT performance is close to the KF baseline on the data association metrics, it actually outperforms the KF model significantly on state estimation. This result also indicates that the S-MOTA metric is useful to distinguish between methods having similar association quality in MOTA results.

To evaluate inference time, we compile the STT model with XLA \cite{sabne2020xla} and run inference on the same scenario as reported in \cite{Sun2022SWFormerSW}. We use a Nvidia PG189 GPU which shares the same hardware architecture as Nvidia T4 GPU but with less memory to meet the power constraints of autonomous vehicles. The inference time for STT alone is $2.9$ ms. Combined with the fastest version of SWFormer as reported in their paper, we can achieve real-time performance for the {end-to-end tracking}.

We also compare our method to TrajectoryFormer~\cite{chen2023trajectoryformer}, which is the current state-of-the-art 3D MOT method on the WOD. We report their CenterPoint~\cite{yin2021center} configuration. It has higher MOTA score than STT due to improved FN (vehicle) and FP (pedestrian) achieved by taking the trajectory hypothesis from track history as model input. We highlight it in a separate row for that a direct comparison with ours is unfair, as TrajectoryFormer uses extra detection boxes. This improvement is orthogonal to our approach. STT still performs better in other two sub-metrics of MOTA. Moreover, TrajectoryFormer does not predict or evaluate on full state estimates, nor does it run in real-time.

% ==========================================

\begin{table*}[!t]
\vspace{3pt}
\centering
\caption{Ablation studies with the proposed STT model on the validation set of Waymo Open Dataset.}
\vspace{-6pt}
\begin{tabular}{@{}l@{\ \ \ \ \ \ }c c c c c c c c c c@{}}
 \toprule
 \multirow{3}{*}{Tracker} & \multirow{3}{*}{Detector} &\multirow{3}{*}{Track Length} &\multirow{3}{3cm}{Joint Optimization w/ State Estimation}  &\multicolumn{2}{c}{Vehicle} &\multicolumn{2}{c}{Pedestrian} \\
 \cmidrule(lr){5-6} \cmidrule(lr){7-8}
 &  & & & MOTA$\uparrow$ & S-MOTA$\uparrow$  &MOTA$\uparrow$ & S-MOTA$\uparrow$\\
 \midrule
\multicolumn{8}{l}{\!\!\!\!\textbf{Joint Optimization of Association and State Estimation}}\\
{\ \ \ \ \ \ }{STT} &SWFormer\cite{Sun2022SWFormerSW}  &10 &N &56.4 &30.9 &55.9 &13.1  \\
{\ \ \ \ \ \ }{STT} &SWFormer\cite{Sun2022SWFormerSW}   &10 &Y &\textbf{58.2} &\textbf{48.0} &\textbf{59.9} &\textbf{55.2}  \\
  \midrule
 \multicolumn{8}{l}{\!\!\!\!\textbf{Long-term Temporal Modeling}}\\
 {\ \ \ \ \ \ }{STT} &SWFormer\cite{Sun2022SWFormerSW}   &3   &Y &58.1 &37.7 &59.9 &52.9  \\
 {\ \ \ \ \ \ }{STT} &SWFormer\cite{Sun2022SWFormerSW}   &5   &Y &\textbf{58.2} &40.4 &\textbf{60.0} &54.1  \\
 {\ \ \ \ \ \ }{STT} &SWFormer\cite{Sun2022SWFormerSW}   &10  &Y &\textbf{58.2} &48.0 &59.9 &55.2  \\
 {\ \ \ \ \ \ }{STT} &SWFormer\cite{Sun2022SWFormerSW}   &20  &Y &\textbf{58.2} &\textbf{49.2} &\textbf{60.0} &\textbf{55.4}  \\
  \midrule
 \multicolumn{8}{l}{\!\!\!\!\textbf{Tracking Performance with Different Detectors}}\\
 {\ \ \ \ \ \ }{Kalman Filter} &UPillar\cite{Leng2022LidarAugmentSF}  &N/A &N/A &55.7 &34.0 &57.1 & 39.8 \\
 {\ \ \ \ \ \ }STT &UPillar\cite{Leng2022LidarAugmentSF}   &10   &Y &\textbf{57.1} &\textbf{46.3}  &\textbf{57.4} &\textbf{52.1}  \\
 {\ \ \ \ \ \ }Kalman Filter &SWFormer\cite{Sun2022SWFormerSW}   &N/A &N/A &56.5 &34.6 &59.7 &41.8  \\
 {\ \ \ \ \ \ }STT &SWFormer\cite{Sun2022SWFormerSW}   &10  &Y &\textbf{58.2} &\textbf{48.0} &\textbf{59.9} &\textbf{55.2}  \\
 \bottomrule
\end{tabular}

\label{table:box-ablation}
\vspace{-20pt} 
\end{table*}

\subsection{MOTP$_\text{S}$ Results}
To further understand the improvements of STT on state estimation, we report the MOTP$_\text{S}$ metric results for STT and two baselines: i) Kalman Filter, and ii) SWFormer+State Head (SH), for which we add a state head to the original SWFormer detector to predict velocity and acceleration for each detected box. The three methods all use the same detection model, which removes the impact of detection quality and allows us to concentrate on the performance of state estimation itself.

As shown in Table~\ref{table:state}, our STT model achieves the best overall state estimation results compared with the two baselines. In terms of velocity estimation, SWFormer+SH is surprisingly the best state estimator for static objects, but STT performs better for moving objects. SWFormer+SH also produces the highest value of $\left|\text{MOTP}_\text{velocity}\right|$ whereas STT has the lowest, indicating that the superior performance of SWFormer+SH on static objects may due to overfitting. On the other hand, the KF baseline struggles to predict accurate states for static objects but can achieve decent performance on moving ones. This may be because small jittering from static objects can create large noise in KF state estimation while learning-based methods are more robust to this. 

The relative gain of STT is more prominent for the acceleration estimation. STT achieves the best acceleration for moving objects and comparable performance with the SWFormer+SH on static objects. STT has the lowest variance compared to the two baselines as reflected by $\left|\text{MOTP}_\text{acceleration}\right|$. Acceleration, as a second order statistic, is more challenging to estimate. Therefore, models must be able to robustly handle small noise and effectively reason about long-term motion. STT possesses both of these qualities, and its robustness and consistency are reflected in the metric results.

\subsection{Ablation Studies}
\noindent{\textbf{Joint optimization with state estimation is important.}} One of the key innovations of STT is its unified learning framework which jointly optimizes for both data association and state estimation tasks. To validate the claim that the joint optimization with state estimation can improve the data association performance, we create a STT baseline that is only trained with the data association loss. The results are reported in the first two rows of Table~\ref{table:box-ablation}. With the joint optimization of state estimation and data association, STT achieves MOTA improvement of +1.8 and +4 for the vehicle and pedestrian classes, respectively. Similarly, S-MOTA improvements of +17.1 and +42.1 are observed for these two classes from STT. These results suggest that data association and state estimation are highly complementary tasks that should be jointly optimized.

\noindent\textbf{Longer-term temporal modeling improves data association quality with more accurate state estimation.} To verify the impact of the temporal features on tracking performance, we evaluate STT with different track history lengths. The results, shown in rows 3 to 6 of Table~\ref{table:box-ablation}, demonstrate that longer track history can lead to improved tracking performance. The MOTA score increases as the track history length increases to 5, after which it saturates. However, the S-MOTA score continues to increase by a large margin, even for track history lengths of 20. This suggests that longer-term temporal modeling is critical for data association and state estimation tasks.

\noindent{\textbf{Improvements from STT are robust with different detectors.}} As our KF baseline experiment shows, the performance of a tracking system can be significantly affected by the quality of the upstream object detector. To understand the sensitivity of STT to different detectors, we compared STT and KF using two different detectors: SWFormer~\cite{Sun2022SWFormerSW} and UPillar~\cite{Leng2022LidarAugmentSF}. The results in Table~\ref{table:box-ablation} show that our STT model outperforms the Kalman Filter on all metrics with different object detectors, which indicates that our model is robust to the choice of detector.

%% file: conclusion.tex
\section{Conclusion}
In this paper, we propose STT, a transformer-based model that jointly conducts data association and state estimation in one model. We emphasize the importance of this joint estimation task for autonomous driving, which requires consistent tracking and accurate state estimation for objects in 3D real-world-space. To address the limitations of existing evaluation methods, we extend MOTA metrics to S-MOTA, which enforces the consideration of state estimation quality when evaluating association quality, and MOTP to $\text{MOTP}_s$, which captures broader motion state of objects. Evaluation has shown that STT achieves the competitive results on the Waymo Open Dataset with strong performance in state estimation. We hope that our proposed solutions and extended metrics will facilitate future work in this area.
\label{sec:conclusion}

%% file: acknowledgement.tex
\noindent\textbf{Acknowledgements.}{ 
We would like to thank Luming Tang, Andy Tsai, Shirley Chung, Yang Wang, Chao Jia, Zhaoqi Leng, Yu Zhu, Nichola Abdo, Henrik Kretzschmar, Marshall Tappen, and Dragomir Anguelov for their invaluable contributions to this paper.}